# Restricted Parallelism in Object-Oriented Lexical Parsing

**Peter Neuhaus**         **Udo Hahn**

Freiburg University

Computational Linguistics Lab

Europaplatz 1, D-79085 Freiburg, Germany

`{neuhaus,hahn}@coling.uni-freiburg.de`

## Abstract

We present an approach to parallel natural language parsing which is based on a concurrent, object-oriented model of computation. A depth-first, yet incomplete parsing algorithm for a dependency grammar is specified and several restrictions on the degree of its parallelization are discussed.

## 1 Introduction

There are several arguments why computational linguists feel attracted by the appeal of parallelism for natural language understanding (for a survey, cf. Hahn & Adriaens (1994)): the ubiquitous requirement of enhanced efficiency of implementations, its inherent potential for fault tolerance and robustness, and a flavor of cognitive plausibility based on psycholinguistic evidences from the architecture of the human language processor. Among the drawbacks of parallel processing one recognizes the danger of greedy resource demands and communication overhead for processors running in parallel as well as the immense complexity of control flow making it hard for humans to properly design and debug parallel programs.

In this paper, we will consider a framework for parallel natural language parsing which summarizes the experiences we have made during the development of a concurrent, object-oriented parser. We started out with a rather liberal conception which allowed for almost unconstrained parallelism. As our work progressed, however, we felt the growing need for restricting its scope as a continuous "domestication process". While still keeping the benefits of parallelism, we have arrived at a point where we argue for a basically serial model patched with several parallel phases rather than a basically parallel model with few synchronization checkpoints. Primarily, this change in perspective was due to large amounts of artificial ambiguities that could be traced to "blind" parallel computations with excessive resource allocation requirements. Continuously taming the parallel activities of the parser and, furthermore, sacrificing highly esteemed theoretical principles such as the completeness of the parser, i.e., the guaranteed production of all analyses for a given input, led us to determine those critical portions of the parsing process which can reasonably be pursued in a parallel manner and thus give real benefits in terms of efficiency and effectiveness.

## 2 Design Requirements for the Parser

The application framework for the parsing device under consideration is set up by the analysis of real-world expository texts (*viz.* information technology product reviews and medical findings reports). The parser operates as the NLP component of a text knowledge acquisition and knowledge base synthesis system.

The analysis of texts (as opposed to sentences in isolation) requires the consideration of discourse phenomena. This includes the interaction of discourse entities (organized in focus spaces and center lists) with structural descriptions from the parser and conceptual information from the domain knowledge base. Thus, different knowledge sources have to be integrated in the course of an incremental text understanding process.

Within realistic NLP scenarios the parsing device will encounter ungrammatical and extragrammatical input. In any of these cases, the parser should guarantee a robust, graceful degradation performance, i.e., produce fragmentary parses and interpretations corresponding to the degree of violation or lack of grammar constraints. Depending on the severity of fragmentation, changes in the parsing strategies which drive the text analysis might also be reasonable.

These requirements obviously put a massive burden on the control mechanisms of a text understanding system. In particular, entirely serial control schemata seem inappropriate, since they would introduce artificial serialization constraints into basically parallel processes (Waltz & Pollack, 1985).

# 3 Object-oriented Lexical Parsing

In this section, we introduce the PARSETALK system, whose specification and implementation is based on an object-oriented, inherently concurrent approach to natural language analysis. We consider constraints which introduce increasing restrictions on the parallel execution of the parsing task. This leads us to a parsing algorithm with restricted parallelism, whose experimental evaluation is briefly summarized.

## 3.1 The PARSETALK Model

The PARSETALK grammar we use (for a survey, cf. Bröker et al. (1994)) is based on binary relations between words, e.g., dependency relations between a head and its modifier, or textual relations between an anaphor and its antecedent. Restrictions on possible relations are attached to the words, e.g., expressed as valencies in the case of dependency relations, yielding a strictly lexicalized grammar in the sense of Schabes et al. (1988). The individual behavior of words is generalized in terms of word classes which are primarily motivated by governability or phrasal distribution; additional criteria include inflection, anaphoric behavior, and possible modifiers. A word class specifies morphosyntactic features, valencies, and allowed orderings for its instances. Further abstraction is achieved by organizing word classes at different levels of specificity in terms of inheritance hierarchies. The specification of binary constraints already provides inherent means for robust analysis, as grammatical functions describe relations between words rather than well-formed constituents. Thus, ill-formed input does often still have an (incomplete) analysis.

The PARSETALK parser (for a survey, cf. Neuhaus & Hahn (1996)) generates dependency structures for sentences and coherence relations at the text level of analysis. In order to establish, e.g., a dependency relation the syntactic and semantic constraints relating to the head and its prospective modifier are checked in tandem. Due to this close coupling of grammatical and conceptual constraints syntactically possible though otherwise disallowed structures are filtered out as early as possible. Also, the provision of conceptual entities which are incrementally generated by the semantic interpretation process supplies the necessary anchoring points for the continuous resolution of textual anaphora and ellipses (Strube & Hahn, 1995; Hahn et al., 1996).

The lexical distribution of grammatical knowledge one finds in many lexicalized grammar formalisms (e.g., LTAGS (Schabes et al., 1988) or HPSG (Pollard & Sag, 1994)) is still constrained to declarative notions. Given that the control flow of text understanding is globally unpredictable and, also, needs to be purposefully adapted to critical states of the analysis (e.g., cases of severe extragrammaticality), we drive lexicalization to its limits in that we also incorporate procedural control knowledge at the lexical grammar level. The specification of lexicalized communication primitives allows heterogeneous and local forms of interaction among (groups of) lexical items. We, nevertheless, take care not to mix up both levels and provide a formally clean specification platform in terms of the actor model of computation (Agha & Hewitt, 1987). In this model each object (*actor*) constitutes a process on its own. Actors communicate by sending messages, usually, in an asynchronous mode. Upon reception of a message, the receiving actor processes the associated method, a program composed of several grammatical predicates, e.g., SYNTAXCHECK, which accounts for morphosyntactic or word order constraints, or CONCEPTCHECK, which refers to the terminological knowledge representation layer and accounts for type and further conceptual admissibility constraints (number restrictions, etc.).

The grammatical description of single words is organized in a hierarchy of so-called *word actors* which not only inherit the declarative portions of grammatical knowledge, but are also supplied with lexicalized procedural knowledge that specifies their parsing behavior in terms of a message protocol. A specialized actor type, called *phrase actor*, comprises word actors which are connected by dependency relations and encapsulates information about that phrase.

## 3.2 Parallelism in Parsing

In the following, we discuss three stages of increasing restrictions of parallelism at the word level, all of which were considered for the design of the algorithm provided in Section 3.3.

**Unbounded Parallelism.** Brute-force parsing models such as the primordial soup algorithm (Janssen et al., 1992), at first sight, exhibit a vast potential for parallel execution, since the central operation of building a structure from two independent parts (in our application, e.g., the combination of a head and a single modifier) apparently does not require any centralized control. In such an entirely unconstrained parallel model, a word actor is instantiated from the input string and sends search messages to *all other* word actors in order to establish a dependency relation, eventually generating a complete parse of the input.

Consider, however, the case in which a Noun is preceded by a Determiner and an Adjective. In order to form the noun phrase [Det Adj [N]] two computation sequences will occur in a primordial soup parser: attaching Det to the N first, then adding Adj, or *vice versa*. Hence, the major drawback of unrestricted parallel algorithms is their non-confluency and, subsequently, either the large (exponential, in the worst case) number of spuriously ambiguous analyses, or the global operation of subsequent duplicate elimination. This led us to restrain from unbounded parallelism and, rather, guarantee confluent behavior by the design of the parsing algorithm.

**Confluency.** In the first prototype, we enforced confluency by an incremental structure-building condition on the basis of a synchronization schema. Messages were forwarded strictly from right to left wandering through the preceding context rather than being broadcasted. Partial structures were organized such that a message which could be successfully processed at a larger structure was not forwarded to any of its constituent parts. But still, the number of ambiguities remained prohibitively large, often due to unnecessary partial structures with large discontinuities. For instance, any determiner preceding a noun forms a new structure, with the Det modifying the N. Usually, a contiguity restriction would filter out those structures given perfectly well-formed input. But such a restriction is detrimental to requirements set up in a realistic text parsing environment, in which the analysis of (possibly large) fractions of un- as well as extragrammatical input must be skipped. Furthermore, ordering restrictions on dependency analyses for German can be formulated more transparently, if discontinuous structures are allowed.

**Depth-First Approach.** These experiences led to a redesign of the first prototype. The parser's forwarding mechanism for search messages was further restricted to circumvent the above mentioned problem of erroneous discontinuous (over)analyses. In this approach, we let phrases that constituted alternative analyses for the *same* part of the input text be encapsulated in a *container actor*. Container actors play a central role in controlling the parsing process, because they encapsulate information about the preceding containers that hold the left context and the chronologically previous containers, i.e., a part of the parse history. Container actors comprising single-word phrases are called *lexical containers*. All phrases in the *active* container send a search message to the current *context* container that forwards them to its encapsulated phrases. The search messages are then asynchronously distributed to words within each phrase. If at least one attachment to one of these phrases is possible, no further forwarding to containers which cover text positions preceding the current position will occur. Thus, the new container composed at this stage will contain only those phrases that were encapsulated in the context container *and* that could be enlarged by attaching a phrase from the active container.

This procedure enforces a *depth-first* style of progression, leaving unconsidered many of the theoretically possible combinations of partial analyses. Still, some information has to be retained in order to backtrack after failures or to employ alternative parsing strategies. We encounter a trade-off between robustness, efficiency, and completeness in parsing. If we were to allow for *unrestricted* backtracking, we would just trade in run-time complexity for space complexity (for a more detailed discussion, cf. Neuhaus & Hahn (1996)). Hence, we rather restrict backtracking to those containers in the parse history which hold governing phrases, while the containers with modifying phrases are immediately deleted after attachment[1].

### 3.3 Restricted Parallel Parsing Algorithm

The parsing algorithm of the PARSETALK system is centered around the head search process of the currently active word actor. If it fails, a modifier search process is triggered; if it succeeds, a new dependency structure is constructed combining the partial analyses. In case both of these protocols are not successful, containers may be skipped so that discontinuous analyses may occur. If the skipping process encounters a linguistically valid boundary (in the most trivial case, the punctuation mark of the previous sentence) it stops and switches to a backtracking mode leading to a kind of roll-back of the parser invalidating the currently pursued analysis. In a companion paper (Neuhaus & Hahn, 1996), we give an integrated description of the various subprotocols needed for head/modifier search, ambiguity handling, skipping, backtracking, preferential and predictive parsing.

In this paper, we concentrate instead on the basic message passing patterns for the establishment of dependency relations, *viz.* the searchHead protocol, and its concurrency aspects. For illustration purposes we here introduce the protocol in a diagrammatic form (Figs. 1 to 3). The figures depict the main steps of word actor initialization, head search, and phrasal attachment. This format eases communication, while formal specifications based on a temporal logic framework are used when dealing with formal properties of the parser (cf., e.g., Schacht (1995) for a partial termination proof of the receipt handler introduced below).

The parser is started by an analyze: message directed to a ParserActor, which is responsible for the global administration of the parsing process (cf. Fig. 1). It instantiates a LexicalContainer-Actor that encapsulates the (potentially) ambiguous readings of the first word of the text, as accessed from the lexicon and corresponding word classes.

Upon receiving the analyzeWithContext: message from the ParserActor (cf. Fig. 2),

---

[1] Hence, the incompleteness property of our parser stems from the selective storage of analyses (i.e., an "incomplete chart" in chart terms), partially compensated by reanalysis.

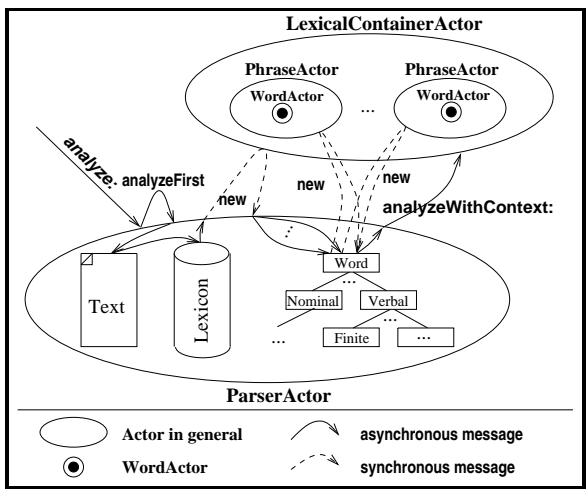

Figure 1: Protocol for Word Actor Initialization

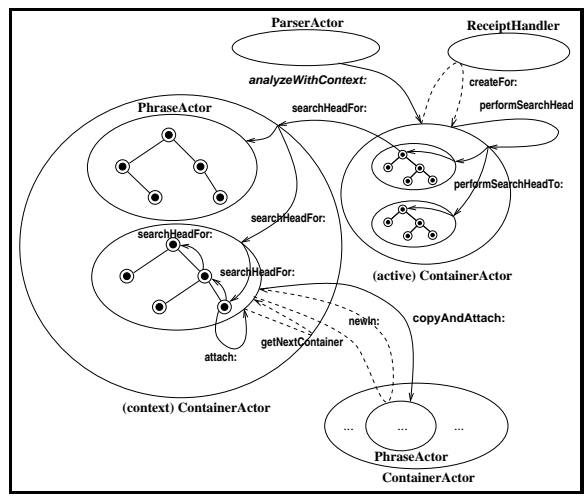

Figure 2: Protocol for the Search for a Head

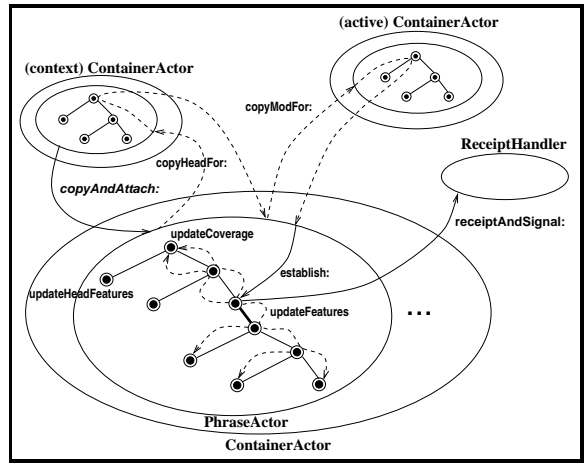

Figure 3: Protocol for Establishing a Dependencies

a `ReceiptHandler` is instantiated by a synchronous message, intended to detect the partial termination of the subsequently started search protocol. The `performSearchHead` message triggers (via `performSearchHeadTo:` messages) asynchronous `searchHeadFor:` messages to be forwarded by each receiving `PhraseActor` to its rightmost `WordActor`. From this origin, the search message can be distributed to all word actors at the right "rim" of the dependency tree by simply forwarding it to the respective heads. After forwarding[2], each `searchHead` message evokes the check of syntactic and semantic restrictions by the corresponding methods. These restrictions must be met in order to establish a dependency relation between the receiving word actor and the sender of the message. Provided that these constraints are fulfilled, an `attach:` message is sent to the encapsulating `PhraseActor`. Before the new composite phrase can be built, the address of the next container actor must be determined. Accordingly, the `getNextContainer` message either returns this address directly or, if it is not available yet, it will *create* the next container actor first (the actual creation protocol is not shown). The `newIn:` message subsequently creates a new `PhraseActor` that will encapsulate the word actors of the new phrase. Notice, that several `attach:` messages can be received by a phrase, because the `searchHead` messages are evaluated in parallel by its word actors.

In order to actually build the new phrase a `copyAndAttach:` message is sent[3]. Fig. 3 depicts the copying of the governing and modifying phrases into the new `PhraseActor` by `copyHeadFor:` and `copyModFor:` messages, respectively — copying enables alternative attachments in the concurrent system, i.e., no destructive operations are carried out. Note that in contrast to the `searchHead` message the `PhraseActor` forwards the copy message to its root actor from where it is distributed in the tree. The dependency relation connecting the copied phrases (indicated by the bold edge in the newly built `PhraseActor`) is created by the `establish:` message. Since word actors hold information (such as features or coverage) locally, updates are necessary to propagate the effects of the creation of the relation. After updating information at relevant word actors in the resultant tree, successful termination of the search message is signalled to the `ReceiptHandler`. If none of the receipts signals success to the handler, the search head protocol will be followed by modifier search or backtracking[4] protocols not shown here (cf.

---

[2] Since forwarded messages are sent asynchronously the processing of the searchHeadFor: message takes place concurrently at the forwarding sender and the respective receivers.

[3] As an alternative to the immediate establishment of a dependency relation, a headFound message can be returned to enable the subsequent selection of preferred attachments (cf. Neuhaus & Hahn (1996) for such a protocol extension).

[4] Thus synchronization of protocols enables word-wise scanning, backtracking, etc. This avoids severe problems usually encountered in parsers with unrestricted parallelism.

Neuhaus & Hahn (1996)). In the scenario we have discussed, the `ReceiptHandler` eventually will detect the success and the termination of the search head protocol. Next, the new `ContainerActor` will be sent an `analyzeWithContext:` message to continue the parsing process.

### 3.4 Preliminary Experimental Evaluation

The efficiency gain that results from the parser design introduced in the previous sections can empirically be demonstrated by a comparison of the PARSETALK system (abbreviated as "PT" below) with a standard active chart parser[5] (abbreviated as "CP"). Since the chart parser is implemented in Smalltalk, while the PARSETALK system is implemented in Actalk (Briot, 1989) — a language which simulates the parallel execution of actors —, a direct comparison of run times is not reasonable (though even at that level of consideration the PARSETALK system outperforms the chart parser). We therefore compare, at a more abstract computation level, the number of method executions given exactly the same dependency grammar[6]. The computationally most expensive methods we consider are SYNTAXCHECK and CONCEPTCHECK (cf. Section 3.1). Especially the latter consumes large computational resources, since for each interpretation variant a knowledge base context has to be built and conceptual consistency must be checked. Therefore, it is only considered when the syntactic criteria are fulfilled. The number of calls to these methods for a sample of 13 randomly chosen, increasingly complex sentences from the information technology domain test library is given by Fig. 4 ("CP.syn" and "PT.syn") and Fig. 5 ("CP.con" and "PT.con"). A reduction by a factor of four to five in the (unweighted) average case can be observed applying the PARSETALK strategy.

Furthermore, the PARSETALK parser, by design, is able to cope with discontinuities stemming from un- or extragrammatical input. The performance of a revised version of the chart parser which also handles these cases is given as "CP.disc.syn/con" in the figures. The missing value for sentence 10 results from the chart parser crashing on this input because of space restrictions of the run-time system (the experiments were conducted on a SPARCstation 10 with 64 MB of main memory). The average reduction in compar-ison with the extended version of the chart parser is about six to nine.

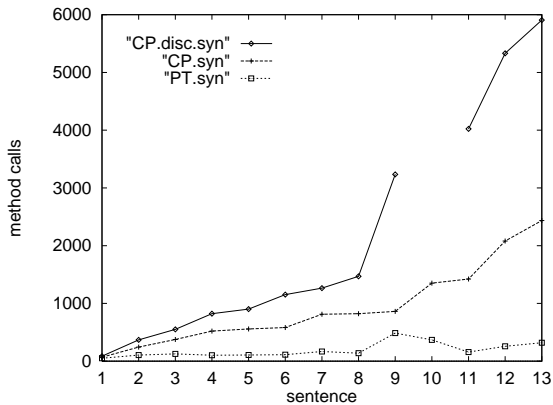

Figure 4: Calls to SYNTAXCHECK

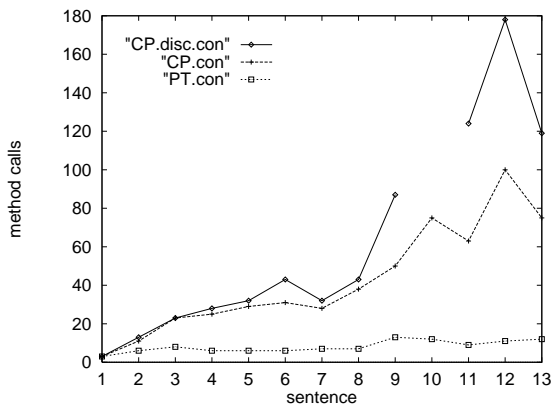

Figure 5: Calls to CONCEPTCHECK

## 4 Related Work

Research on object-oriented natural language parsing actually started with the work of Small & Rieger (1982) on word experts. Based on a conceptual parsing model, this approach took a radical position on full lexicalization and communication based on a strict message protocol. Major drawbacks concerned an overstatement of the role of lexical idiosyncrasies and the lack of grammatical abstraction and formalization. Preserving the strengths of this approach (lexicalized control), but at the same time reconciling it with current standards of lexicalized grammar specification, the PARSETALK system can be considered a unifying approach which combines procedural and declarative specifications at the grammar level in a formally disciplined way. This also distinguishes our approach from another major stream of object-oriented natural language parsing which is almost entirely concerned with implementational aspects of object-oriented programming, e.g., Habert (1991), Lin (1993) or Yonezawa & Ohsawa (1994).

The reasons why we diverge from conventional parsing methodologies, e.g., chart parsing based on

---

[5] Winograd's (1983) chart parser was adapted to parsing a dependency grammar. No packing or structure-sharing techniques could be used, since semantic interpretation occurs online, thus requiring continuous referential instantiation of structural linguistic items (cf. also Section 4).

[6] This can only serve as a rough estimate, since it does not take into account the exploitation of PARSETALK's concurrency. Furthermore, the chart parser performs an extremely resource-intensive subsumption checking method unnecessary in the PARSETALK system.

Earley- or Tomita-style algorithms, are two-fold. First, at the syntactic level, any kind of chart parsing algorithm faces combinatorial problems with non-contiguous grammar specifications (accounting for discontinuous language structures) and, in particular, extra- and ungrammatical language input (cf., e.g., Magerman & Weir (1992) for probabilistic and Lee et al. (1995) for symbolic heuristics to cope with that problem). Thus, under realistic conditions, these techniques loose a lot of their theoretical appeal and compete with other approaches merely on the basis of performance measurements. Second, including semantic considerations, even if we assume efficient syntactic processing for the sake of argument, the question arises how semantic interpretations can be processed in an incremental, comparably efficient way. Though experiments have been run with packing feature structures and interleaving syntactic and semantic analyses (Dowding et al., 1994), or with the intentional underspecification of logical forms (leaving scope ambiguities of quantifiers and negations underdetermined; cf., e.g., Hobbs (1983) or Reyle (1995)), no conclusive evidences have been generated so far in favor of a general method for efficient, online semantic interpretation. As we are faced, however, with the problem to work out text interpretations incrementally and within reasonable resource bounds, we opt for a methodology that constrains the amount of ambiguous structures right at the source. Hence, the incompleteness of the algorithm trades theoretical purism for feasibility of realistic NLP.

## 5 Conclusions

We have presented a restricted approach to parallelism for object-oriented lexicalized parsing. Given the complex control structure requirements of a realistic text understanding system (integrated, incremental, robust processing), we argued for a unifying approach in which declarative grammar constraints are lexically encoded and procedural knowledge can be specified by distinguished lexicalized communication primitives (*viz.* a message passing protocol). This led us to the description of a concurrent parsing algorithm which is characterized by a depth-first, robust, yet incomplete analysis of textual input. We also argued in favor of incompleteness in order to break the text parsing complexity barrier. As a consequence, we do not only supply an efficient parsing procedure but also one that is effective in the sense that it guarantees the generation of conceptual representations of the content of the text under feasible resource demands.

**Acknowledgments.** P. Neuhaus is supported by a grant from DFG within the Freiburg University Graduate Program on *"Human and Artificial Intelligence"*.